\documentclass{llncs}

\usepackage{ mathptmx}
\usepackage{multirow}
\usepackage{amsmath,bm}
\usepackage{graphicx}
\usepackage{subfigure}
\usepackage{array}
\usepackage{makeidx}  
\usepackage{expl3}
\usepackage{url}

\ExplSyntaxOn
\newcommand\latinabbrev[1]{
  \peek_meaning:NTF . {
    #1\@}%
  { \peek_catcode:NTF a {
      #1.\@ }%
    {#1.\@}}}
\ExplSyntaxOff

\def\etal{\latinabbrev{et al}}

\begin{document}

%

%
\mainmatter              
\title{Medical Image Synthesis with Context-Aware Generative Adversarial Networks}
%
%
\author{Dong Nie\inst{1}\thanks{The authors contributed equally to this work}  \and Roger Trullo\inst{2}\footnotemark[1] \and
Caroline Petitjean\inst{2} \and Su Ruan\inst{2} \and
Dinggang Shen\inst{1}\thanks{Corresponding author}}
%
%
\tocauthor{Ivar Ekeland, Roger Temam, Jeffrey Dean, David Grove, Craig Chambers, Kim B. Bruce, and Elisa Bertino}
\institute{University of North Carolina at Chapel Hill, USA\\
\and
Normandie Univ, UNIROUEN, UNIHAVRE, INSA Rouen, LITIS, 76000 Rouen, France}

\maketitle              

\begin{abstract}
Computed tomography (CT) is critical for various clinical applications, e.g., radiotherapy treatment planning and also PET attenuation correction. However, CT exposes radiation during acquisition, which may cause side effects to patients. Compared to CT, magnetic resonance imaging (MRI) is much safer and does not involve any radiations. Therefore, recently, researchers are greatly motivated to estimate CT image from its corresponding MR image of the same subject for the case of radiotherapy planning. In this paper, we propose a data-driven approach to address this challenging problem. Specifically, we train a fully convolutional network to generate CT given an MR image. To better model the nonlinear relationship from MRI to CT and to produce more realistic images, we propose to use the adversarial training strategy and an image gradient difference loss function. We further apply AutoContext Model to implement a context-aware generative adversarial network.
Experimental results show that our method is accurate and robust for predicting CT images from MRI images, and also outperforms three state-of-the-art methods under comparison.
\keywords{Generative Models, GAN, Image Synthesis, Deep Learning, AutoContext}
\end{abstract}
\section{Introduction}
Computed Tomography (CT) imaging is widely used for both diagnostic and therapeutic purposes in various clinical applications. In the cancer radiation therapy, CT image provides Hounsfield units, which is essential for dose calculation in treatment planning. Besides, CT image is also of great importance for attenuation correction of positron emission tomography (PET) in the popular PET-CT scanner \cite{kinahan1998attenuation}.

However, patients are exposed to radiation during CT imaging, which can damage normal body cells and further increase potential risks of cancer. Brenner \etal  \cite{brenner2007computed} reported that 0.4\% of cancers were due to CT scanning performed in the past, and this rate will increase to as high as 1.5 to 2\% in the future. Therefore, the use of CT scan should be done with great caution. Magnetic Resonance Imaging (MRI) on the other hand, is a safe imaging protocol which also provides more anatomical details than CT for diagnostic purposes, but unfortunately cannot be used for either dose calculation or attenuation correction. To reduce unnecessary imaging dose for patients, it is clinically desired to estimate CT images from MR images in many applications.
\begin{figure}[t!]
\centering
  \includegraphics[width=0.5\linewidth]{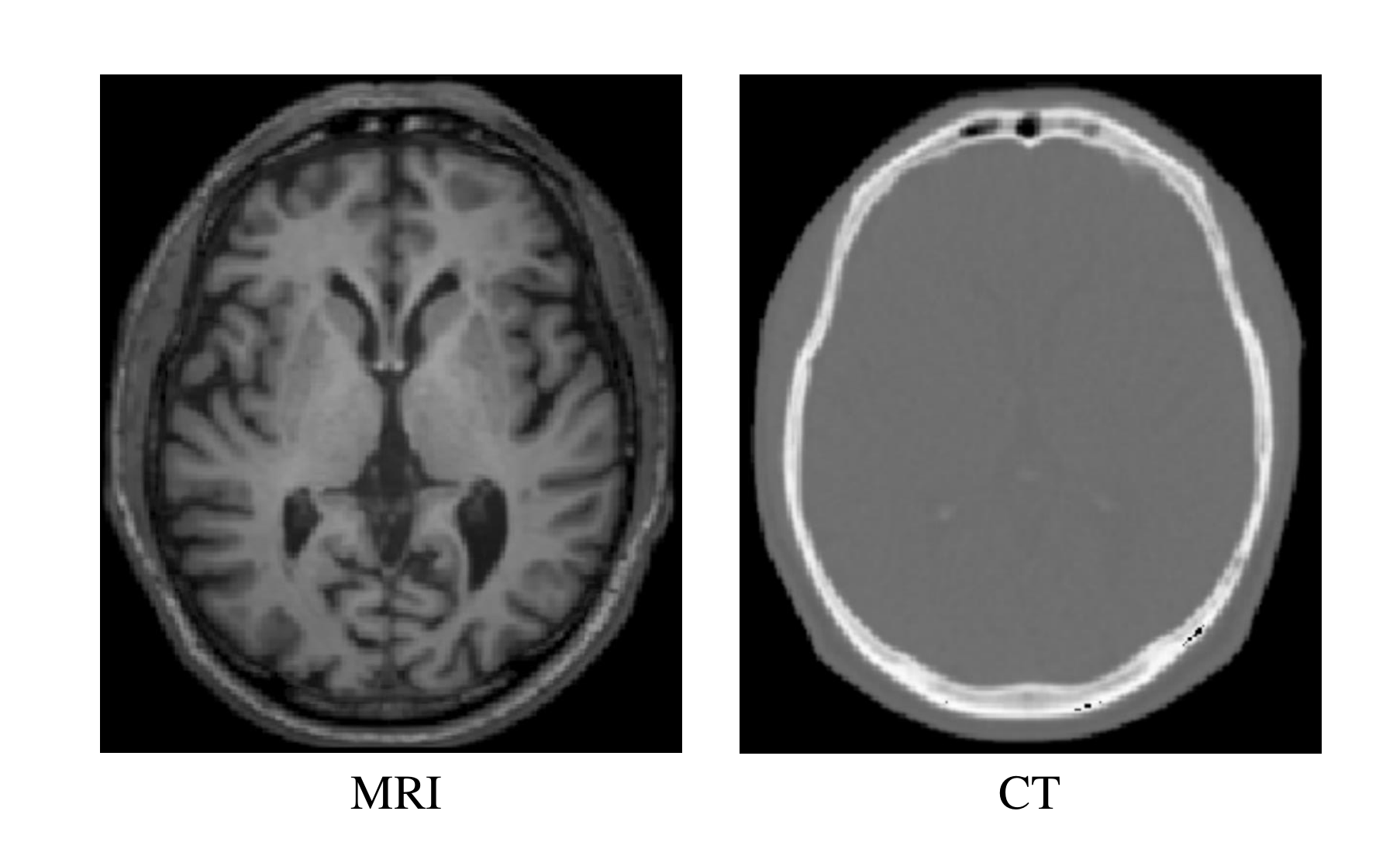}\\
  \caption{A pair of corresponding brain MR (left) and CT (right) images from the same subject}
\label{fig:brain}
\vspace{-5pt}
\end{figure}

It is technically difficult to directly estimate CT image from MR image. As shown in Fig. \ref{fig:brain},  CT and MR images have very different appearances. MR images contain much richer texture information than CT images. Therefore, it is challenging to directly estimate a mapping from MRI to CT.

Recently, many researches focus on estimating one modal image from another modality image, e.g., estimating CT image using MRI data. The first category of methods is image segmentation based methods. Zaidi \etal ~\cite{zaidi2003magnetic} developed a fuzzy clustering technique to segment MR images into different tissue classes, and then refined segmentation manually. Berker \etal ~\cite{berker2012mri} proposed four class tissue segmentation technique with a combined ultrashort-echo-time/Dixon MRI sequence to fulfill MRI-based attenuation correction. These methods segment MR images into different tissue classes, and then assign each class with a known attenuation property. This category of methods highly depends on the segmentation accuracy and always needs manual work to get final accurate results. 
The second category of methods is atlas-based methods. These methods first register an atlas (with the attenuation map) to the new subject's MR image, and then warp the corresponding attenuation map of the atlas to the new MR image as its estimated attenuation map~\cite{catana2010toward}. However, the performance of these atlas-based methods highly depends on the registration accuracy. 
The third category is learning-based methods, in which a non-linear mapping model is learnt from MRI to CT image. Jog \etal ~\cite{jog2014improving} learned nonlinear regression using random forest to improve MR Resolution. Tri \etal ~\cite{huynh2015estimating} presented an approach to predict CT image from MRI using structured random forest.
Since estimating MRI from CT is not a one-to-one mapping task, using only the intensity values cannot effectively distinguish the structural details.
Such methods often have to first represent the input MR image by features and then map them to output the CT image. Thus, the performance of these methods is bound to the quality of the extracted features and how well they can represent the natural properties of the MR image.

On the other hand, recently the convolutional neural network (CNN) ~\cite{lecun2015deep} became popular in both computer vision and medical imaging fields. As a multi-layer and fully trainable model, CNN is able to capture the complex non-linear mapping from the input space to the output space. For the case of 2D images, 2D CNN has been widely used in many applications. However, it is unreasonable to apply 2D CNN to process 3D medical images because 2D CNN considers the image appearance slice by slice, thus potentially causing discontinuous prediction results across slices. To address this issue, 3D CNNs have been proposed.  Ji \etal ~\cite{ji20133d} presented a 3D CNN model for action recognition in an uncontrolled environment. Tran \etal ~\cite{tran2014learning} used 3D CNN to effectively learn spatio-temporal features on a large-scale video dataset. Dong \etal ~\cite{dong2014learning} proposed a deep learning method for single image super-resolution. Li \etal ~\cite{li2014deep} applied deep learning models to estimate the missing PET image from the MR image of the same subject. 

Typically an L2 distance is used as loss function with the assumption that the data is drawn from a Gaussian distribution. This can be problematic in the case of multimodal distributions and tends to produce blurry results in the output images \cite{mathieu2015deep}.
Moreover, recently in the field of image generation, Generative Adversarial Networks (GAN) have achieved state of the art results producing very realistic images in an unsupervised setting \cite{goodfellow2014generative}, \cite{radford2015unsupervised}.

In this paper, we propose to learn the non-linear mapping from MR to CT images through a 3D fully convolutional neural network (FCN), which is a variation of the conventional CNN. Compared to CNN, FCN generates the structured output, which can better preserve the neighborhood information in the predicted CT image. This 3D FCN is used as generator in a Generative Adversarial framework where an adversarial loss term in addition to the conventional reconstruction error with the objective of producing more realistic CT data. The network is trained on patches which makes its view to be restricted to the patch itself and thus cannot provide long-range information. We use Auto-Context Model (ACM) where each stage is trained using the GAN framework for making it context-aware. To the best of our knowledge, this is the first application of the GAN framework in the field of synthetic medical image generation.

The proposed method is evaluated on two real CT/MR datasets. Experimental results demonstrate that our method can effectively predict CT image from MR image, and also outperforms three state-of-the-art methods under the comparison.

\section{Methods}
\label{sec:method}
Deep learning models can learn a hierarchy of features, i.e., high-level features built upon low-level features. CNN~\cite{glorot2010understanding,lecun2015deep} is one popular type of deep learning models, in which trainable filters and local neighborhood pooling operations are applied in an alternating sequence starting with the raw input images. When trained with appropriate regularization, CNN can achieve superior performance on visual object recognition and image classification tasks~\cite{lecun1998gradient}.  However, most of CNNs are designed for 2D natural images. They are not well suited for medical image analysis, since most of medical images are 3D volumetric images, such as MRI, CT and PET. Additionally, the output of a conventional CNN is a single target value, which means that for a full volume, it would be necessary to compute one output for every voxel. This framework has the obvious disadvantage of requiring a large computational time, but also is unable to preserve neighborhood information in the output space. The task of image synthesis can be seen as a regression problem, where for every voxel in the input, an estimated output is required. Typically an Euclidean loss function is used during training for regression networks, which could potentially generate blurry images.

To address the above mentioned problems, we propose a generative adversarial network in which fully convolutional networks form the generator. First, we propose a basic 3D FCN structure to estimate the CT from MRI images. 3D operations can better model the 3D spatial information and thus could solve the discontinuity problem across slices, which can be present in 2D CNN. Second, we utilize the adversarial training strategy \cite{goodfellow2014generative} for the designed network where an additional discriminator network can urge the generator's output to look like real CT as much as possible. We add an image gradient difference term to the loss function of the generator, with the aim of retaining the sharpness of the generated CT. Finally, we employ the Auto-Context model to iteratively refine the output of the generator.

At testing stage, an input MR image is first partitioned into overlapping patches. For each patch, the generator is used to predict the corresponding CT patch. Finally, all predicted CT patches are merged into a single CT image by averaging the intensities of overlapping CT regions.

In the following paragraphs, we will describe the framework that we use  in the MRI-to-CT prediction.

\subsection{Proposed Generative Adversarial Networks}
\label{subsec:gan}
Generative Adversarial Networks (GANs) have been successfully applied in the context of generative models. They were proposed in~\cite{goodfellow2014generative}, and have achieved impressive results in the context of image generation. GANs work by training two different networks: a generator network $G$, and a discriminator network $D$. $G$ is typically an FCN which generates images $G(\bm z)$  from a random noise vector $\bm z$, and $D$ is a CNN which estimates the probability that an input image $x$ is drawn from the distribution of real images; that is, it can classify an input image as real or synthetic. Both networks are trained simultaneously with $D$ trying to correctly discriminate between real and synthetic data, while $G$ is trying to produce realistic images that will confuse $D$. More formally, for $D$ we would like to find its parameters:

\begin{equation}
\max_{D} ~\log(D(x))+\log(1-D(G(\bm z)))
\label{eq:obj_D}
\end{equation}
For $G$ we would like to optimize
\begin{equation}
\max_{G} ~\log(D(G(\bm z)))
\label{eq:obj_G}
\end{equation}

\begin{figure}
  \includegraphics[width=1\linewidth]{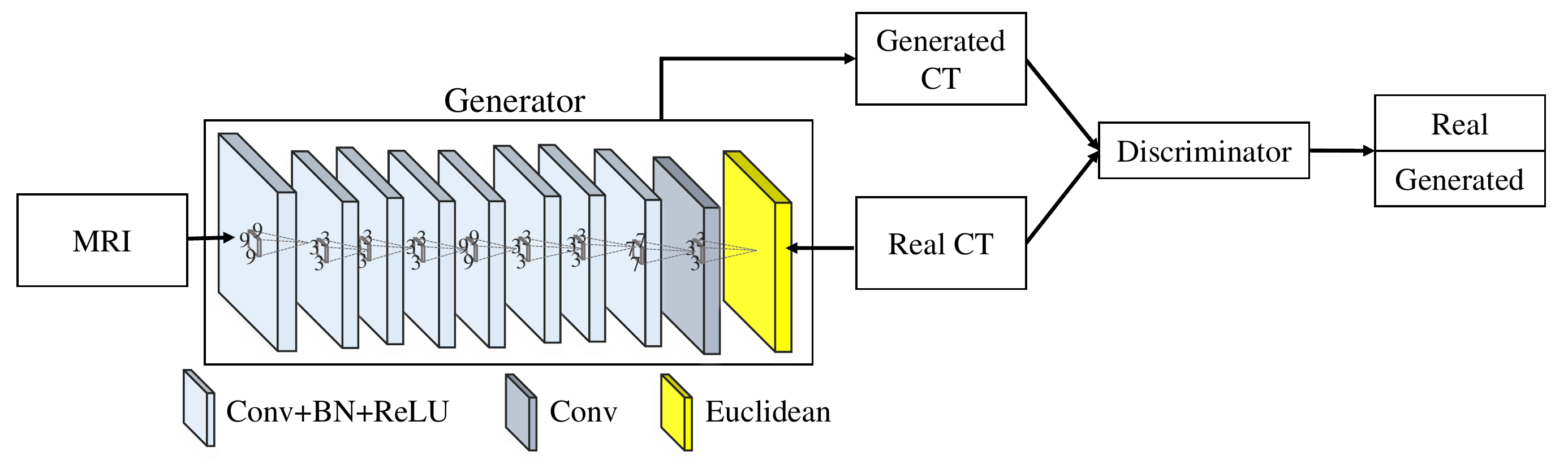}
  \caption{Architecture used in the Generative Adversarial setting used for estimation of synthetic images.}
  \label{fig:gan} 
\end{figure}

Inspired by the  work in \cite{mathieu2015deep}, where the authors use the GAN framework for video frame prediction using the input frames as input instead of a random vector, we design the GAN framework shown in Fig.~\ref{fig:gan}. Our network includes a generator which aims to estimate the CT and a discriminator which aims to distinguish the real CT from the generated one. Specifically, we minimize the binary cross entropy (bce) between the decisions of $D$ and the correct label (real or synthetic), while the network $G$ is trying to minimize the binary cross entropy between the decision done by $D$ and the wrong label for the generated images, in addition to the traditional reconstruction error. In this way, $D$ is trying to distinguish between real CT data, and the CT data generated by $G$. At the same time, $G$ is trying to produce more realistic CT images such that $D$ gets completely confused and cannot perform better than chance.

Concretely, the loss function for $D$ is defined as:

\begin{equation}
L_{D}=L_{bce}(D(x),1)+L_{bce}(D(G(x)),0)
\label{eq:loss_D}
\end{equation}
where $x$ is the input MRI image and
\begin{equation}
L_{bce}(\hat{y},y)=-\frac{1}{N}\sum_{i=1}^{N}{y_i~log\hat{y_i}+(1-y_i)~log(1-\hat{y_i})}
\label{eq:bce}
\end{equation}
$N$ is the number of samples in the minibatch, ~$y \in \{0,1\}$ represents the label of the input data (0 for the generated and 1 for the real CT), and $\hat{y} \in [0,1]$ is the estimated label by the discriminator network.

In the case of $G$, as mentioned above, we use a loss that includes an adversarial term and a reconstruction error. It is defined as:

\begin{equation}
L_{G}=\lambda_1L_{bce}(D(G(x)),1)+\lambda_2\lVert Y-G(x) \rVert_2^{2}
\label{eq:loss_G}
\end{equation}
where $Y$ is the  corresponding ground-truth CT.

Similar to \cite{mathieu2015deep}, in order to deal with the inherently blurry predictions obtained from the L2 loss function (Eq.\ref{eq:loss_G}), we propose to embed an image gradient difference loss function in the generator training. The gradient difference loss (gdl) between the generated CT and the real CT is given by:

\begin{equation}
L_{gdl} = \left| {\left| {\nabla {Y_x}} \right| - \left| {\nabla {{\widehat Y}_x}} \right|} \right|^2 + \left| {\left| {\nabla {Y_y}} \right| - \left| {\nabla {{\widehat Y}_y}} \right|} \right|^2 + \left| {\left| {\nabla {Y_z}} \right| - \left| {\nabla {{\widehat Y}_z}} \right|} \right|^2
\label{eq:gdl}
\end{equation}
where $Y$ is the ground truth image data, and $\hat{Y}$ is the estimated  by the network. This loss function tries to minimize the difference of the  magnitude of the gradient between the estimated image and the ground-truth CT. In this way, the estimated data will try to keep zones with strong gradients (for example edges) for effectively compensating for the L2 term. This is approximated as finite differences during the implementation. Finally, the total loss used for training the generator $G$ is defined as the weighted sum of all the terms as shown in Eq.\ref{eq:totalLoss}.

\begin{equation}
\label{eq:totalLoss}
L\left( {X,Y} \right) = {L_G}\left( {X,Y} \right) + {\lambda _3}{L_{gdl}}\left( {X,Y} \right)
\end{equation}

The training is performed in an alternating fashion; first, $D$ is updated by taking a minibatch of real CT data and a mini-batch of generated (the output of $G$) data. Then, $G$ is updated by using another mini-batch of samples including MRI and their corresponding CT.

Regarding the architecture, we make use of batch normalization, and also we avoid the use of pooling layers since they reduce the spatial resolution of feature maps. Although this property is desirable for some tasks, such as image classification where the pooling over local neighborhoods can enhance the invariance to certain image distortions, it is not desired in the task of image prediction, where subtle image distortions need to be precisely captured in the prediction process. These properties agree with several works that have extended GAN, showing that some architectural constraints are necessary in order to achieve a stable training. The authors in~\cite{radford2015unsupervised}, for example, were able to obtain very realistic images by using FCN without max pooling and with batch normalization across different layers in both $G$ and $D$ in a  way similar to what we propose.

In Fig.~\ref{fig:gan}, we also show the architecture of our generator network $G$  which has the constraints mentioned above, where the numbers indicate the filter sizes. The network takes as input an MRI image, and tries to generate the corresponding CT image. It has 8 stages containing convolutions, Batch Normalization and ReLu operations with number of filters 32, 32, 32, 64, 64, 64, 32, 32, respectively. The last layer only includes 32 convolutional filters, and its output is considered as the estimated CT. Finally, the Discriminator is a typical CNN architecture including three stages of convolutions+Batch Normalization+ReLu+Max Pooling and a combination of convolution with three fully connected layers, where the first two use ReLu as activation function, and the last one uses sigmoid, whose output represents the likelihood that the input data was drawn from the distribution of real CT. The filter sizes are $5\times 5\times 5$, the numbers of filters are 32, 64, 128 and 256 for the convolutional layers, and the numbers of output nodes in the fully connected layers are 512, 128 and 1.

 

\subsection{Auto-Context Model (ACM) for Refinement}
\label{sec:acm}
\begin{figure}
  \includegraphics[width=0.8\linewidth]{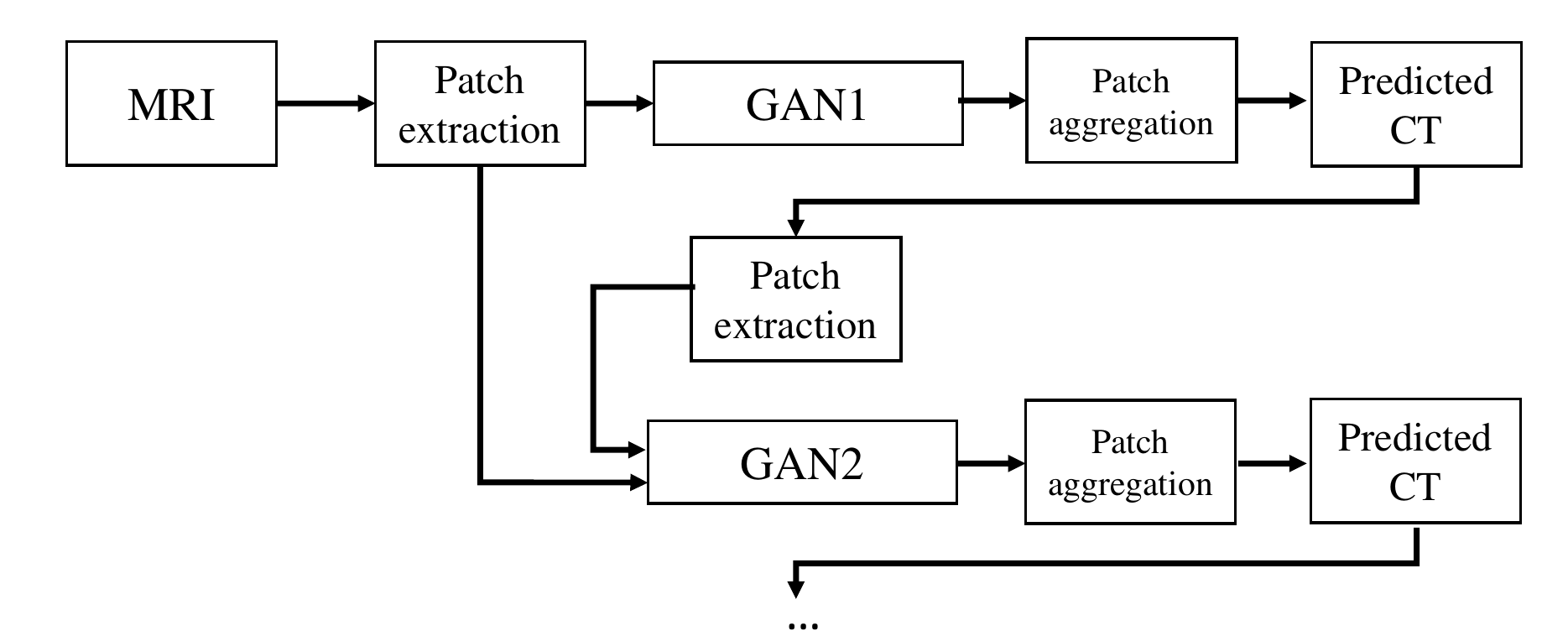}
  \caption{Proposed architecture for ACM with GAN.}
  \label{fig:acm} 
\end{figure}

Since our work is patch-based, the context information available for each training sample is limited inside of the patch. This affects the modeling capacity of our network. One way to enlarge the context during training is by using the Auto-Context model which is commonly used in medical image analysis applications and it has been shown to be very effective. It was first introduced by ~\cite{tu2010auto}, where it was used in the context of segmentation. The idea is to train several classifiers iteratively where each classifier is trained not only with the feature data, but also with the probability map obtained from the previous classifier, which gives to the classifier additional context information. At testing time, the features will be processed for each classifier one after the other, concatenating the probability map to the input features.
This technique has shown to work remarkably well in the task of semantic segmentation of MRI data ~\cite{wang2015links}, where Random Forest was used as classifier. 
In this work, we show that the ACM can be successfully applied also in regression tasks. Since the context features are extracted from the whole previously estimated CT, they can encode information that is not available using only patch based features. In our work, instead of building several classifiers, we use Generator networks, and also instead of concatenating probability maps, we concatenate the output of the previous generator network. Specifically, we train iteratively several GANs that take as input MRI patches and estimate corresponding CT patches. These patches are concatenated as a second channel in the MRI patches, and this new data is used as input during the training of the next GAN. An illustration of this scheme is shown in Fig.\ref{fig:acm}.

\section{Datasets}
We use two datasets to test our proposed methods.
\begin{itemize} 

\item The brain dataset was acquired from 16 subjects
with both MRI and CT scans in the Alzheimer's Disease
Neuroimaging Initiative (ADNI) database (see
\url{www.adni-info.org} for details). The MR images were
acquired using a Siemens Triotim scanner, with voxel
size $1.2\times1.2\times1~\text{mm}^3$ , TE 2.95 ~\text{ms}, TR 2300 ~\text{ms}, and
flip angle $9^\circ$; The CT images, with voxel
size $0.59\times0.59\times3~\text{mm}^3$, were acquired
on a Siemens Somatom scanner. A typical example of preprocessed CT and MR images is given in Fig.~\ref{fig:brain}.

\item Our pelvic dataset consists of 22 subjects, each with MR and CT images. The spacings of CT and MR images are $1.172\times1.172\times1~\text{mm}^3$ and $1\times1\times1~\text{mm}^3$, respectively. In the training stage, CT image is manually aligned to MR image to build the voxel-level correspondence. After alignment, CT and MR images of the same patient have the same image size and spacing. Since only pelvic regions are concerned, we further crop the aligned CT and MR images to reduce the computational burden. Finally, each preprocessed image has a size of $153\times193\times50$ and a spacing of $1\times1\times1~\text{mm}^3$. 
\end{itemize}

\section{Experimental Results}
We extracted randomly MRI patches of size $32\times 32 \times 32$ with corresponding CT of size $16\times 16 \times 16$ using the same center point as training samples. The networks were trained using the Adam optimizer with a learning rate of $10^{-6}$, $\beta_1=0.5$ as suggested in \cite{radford2015unsupervised}, and mini-batch size of 10. The generator was trained using $\lambda_1=0.5, \lambda_2=\lambda_3=1$.

The code is implemented using the TensorFlow library, and it will be publicly released upon acceptance.
To demonstrate the advantage of the proposed method in terms of prediction accuracy, we compare it with three widely used approaches: an atlas-based method, a sparse representation based method, and structured random forest with auto-context model. We used our own implementation of the first two methods, while for the the structured random forest we show the results declared in \cite{huynh2015estimating}. And all experiments are done in a leave-one-out fashion.
\begin{itemize}

\item Atlas-based method ({\bf Atlas}): Here, the MR image of each atlas is first aligned~\cite{vercauteren2009diffeomorphic} onto the target MR image, and the resulting deformation field is used to warp the CT image of each atlas. The final prediction is obtained by averaging all warped CT images of all atlases.

\item Sparse representation based method ({\bf SR}): After warping the atlases to the target image space as described above, a local sparse representation is then performed. 

\item Structured random forest and auto-context model ({\bf SRF+}) based method: Besides the structured random forest, auto-context model (ACM)~\cite{tu2010auto} is further used to iteratively refine the prediction of CT images.

\end{itemize}

To compare the performance of different methods, we utilize mean absolute error (MAE) and peak signal-to-noise ratio (PSNR) as measurements.

\subsection{Impact of Proposed GAN model}
\begin{figure}[t!]
\centering
  \includegraphics[width=0.9\linewidth]{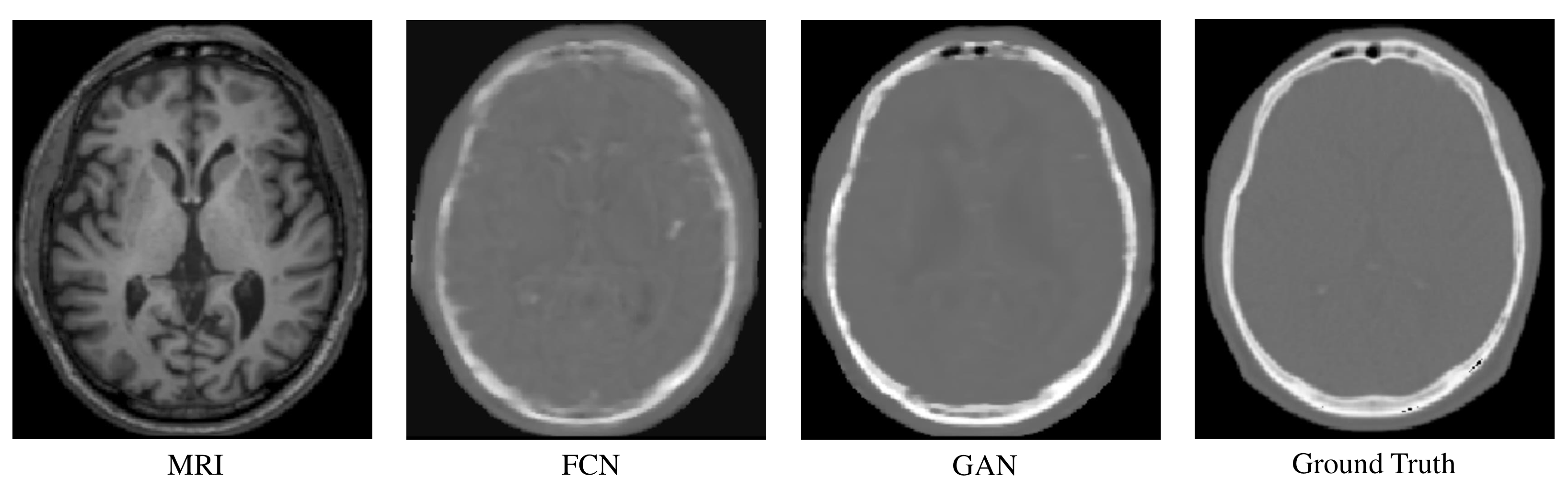}\\
  \caption{Visual comparison for impact of adversarial training. FCN means without adversarial training, and GAN means with adversarial training.}
\label{fig:ganvsfcn}
\vspace{-5pt}
\end{figure}

To show the contribution of the proposed GAN model, we conduct comparison experiments between the traditional FCN (just the generator shown in Fig.~\ref{fig:gan}) and the proposed GAN model. The PSNR are 24.7 and 25.9 for the traditional FCN and the proposed approach, respectively. These results do not include the ACM. We can visualize the results in Fig.~\ref{fig:ganvsfcn}, where the leftmost image is the input MRI, and the rightmost image is the ground-truth CT. We can clearly see that the generated data using the GAN approach has less artifacts than the traditional FCN, by estimating an image that is closer to the desired output not only quantitatively but also qualitatively.

\subsection{Impact of Auto-Context Refinement}
\begin{figure}
\begin{minipage}[t]{0.5\linewidth}
\centering
\includegraphics[width=2.2in]{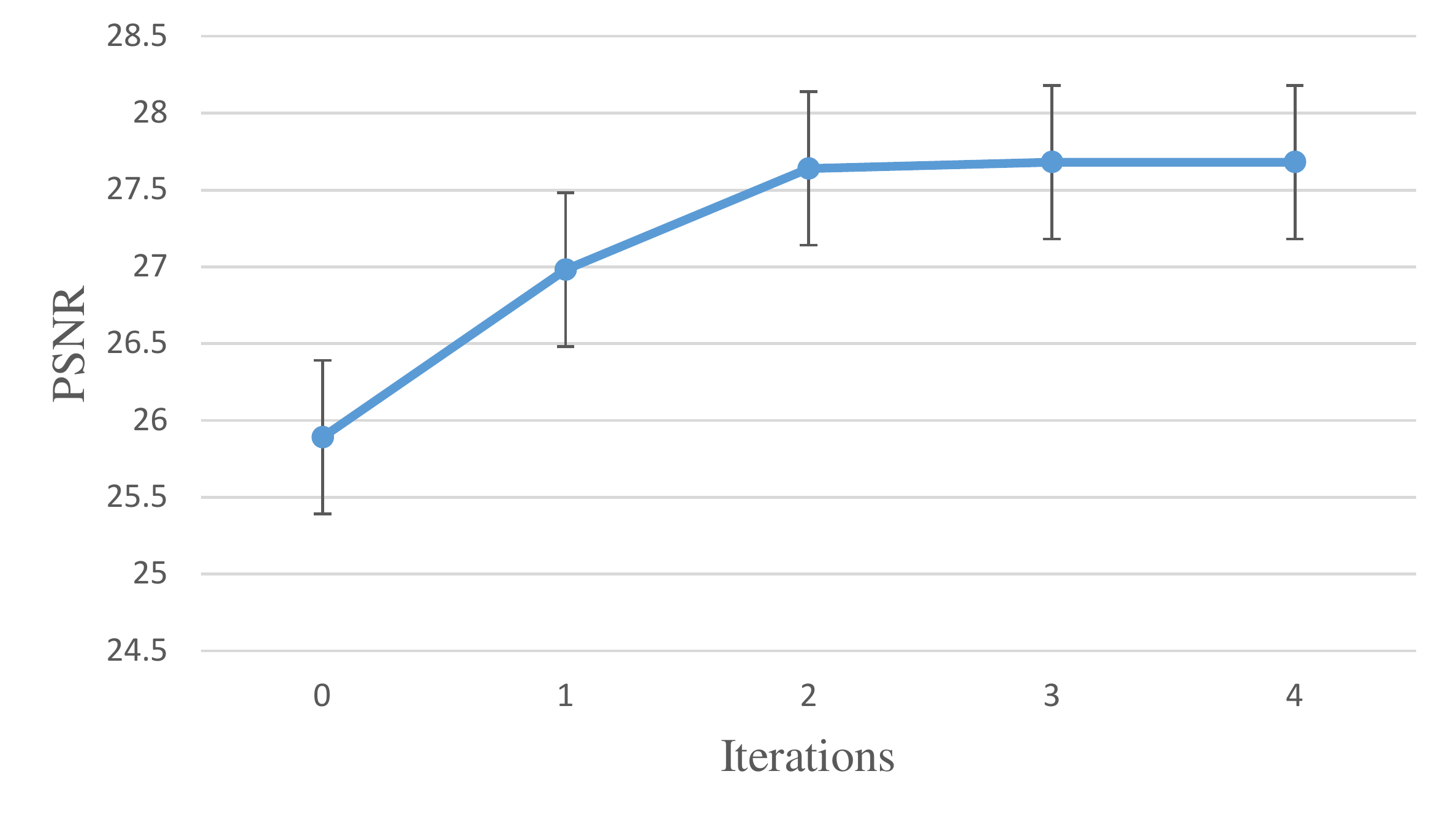}
\label{fig:side:a}
\end{minipage}%
\begin{minipage}[t]{0.5\linewidth}
\centering
\includegraphics[width=2.2in]{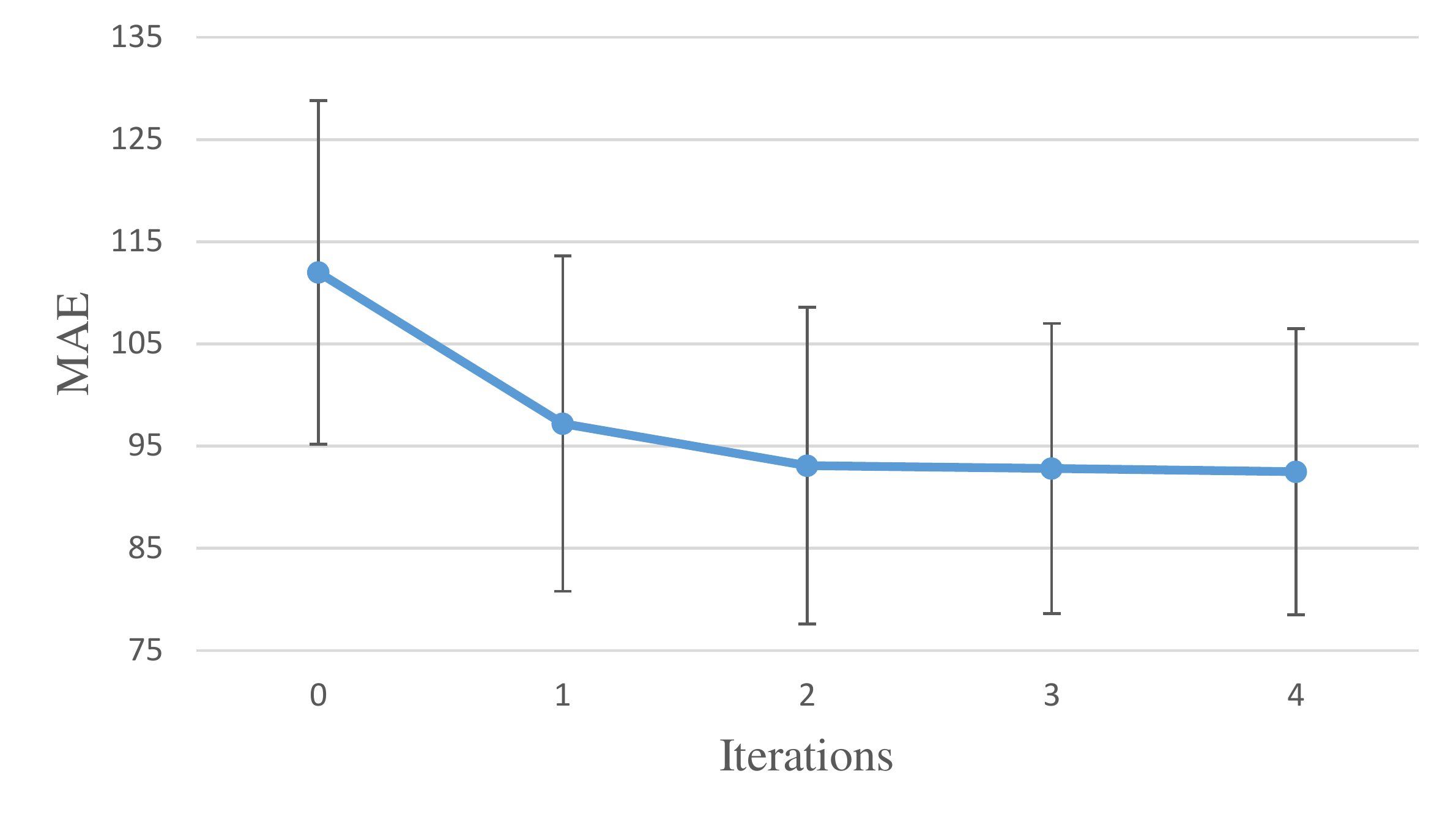}
\label{fig:side:b}
\end{minipage}
\caption{Performance of ACM on brain dataset with iterations.}
\label{fig:acm-iter}
\end{figure}

To show the contribution of ACM, we present the performance of the proposed method with iterations of ACM in Fig.~\ref{fig:acm-iter}. We can quantitatively observe that both MAE and PSNR are improved gradually and consistently with iterations, especially in the first two iterations. This is because ACM could solve the short-range dependency by providing long-range information. Considering the trade-off between benefit and training time, we choose 2 iterations of ACM in our experiments for both datasets.

\subsection{Experimental Results for Brain Dataset}
\begin{figure}[t!]
\centering
  \includegraphics[width=0.9\linewidth]{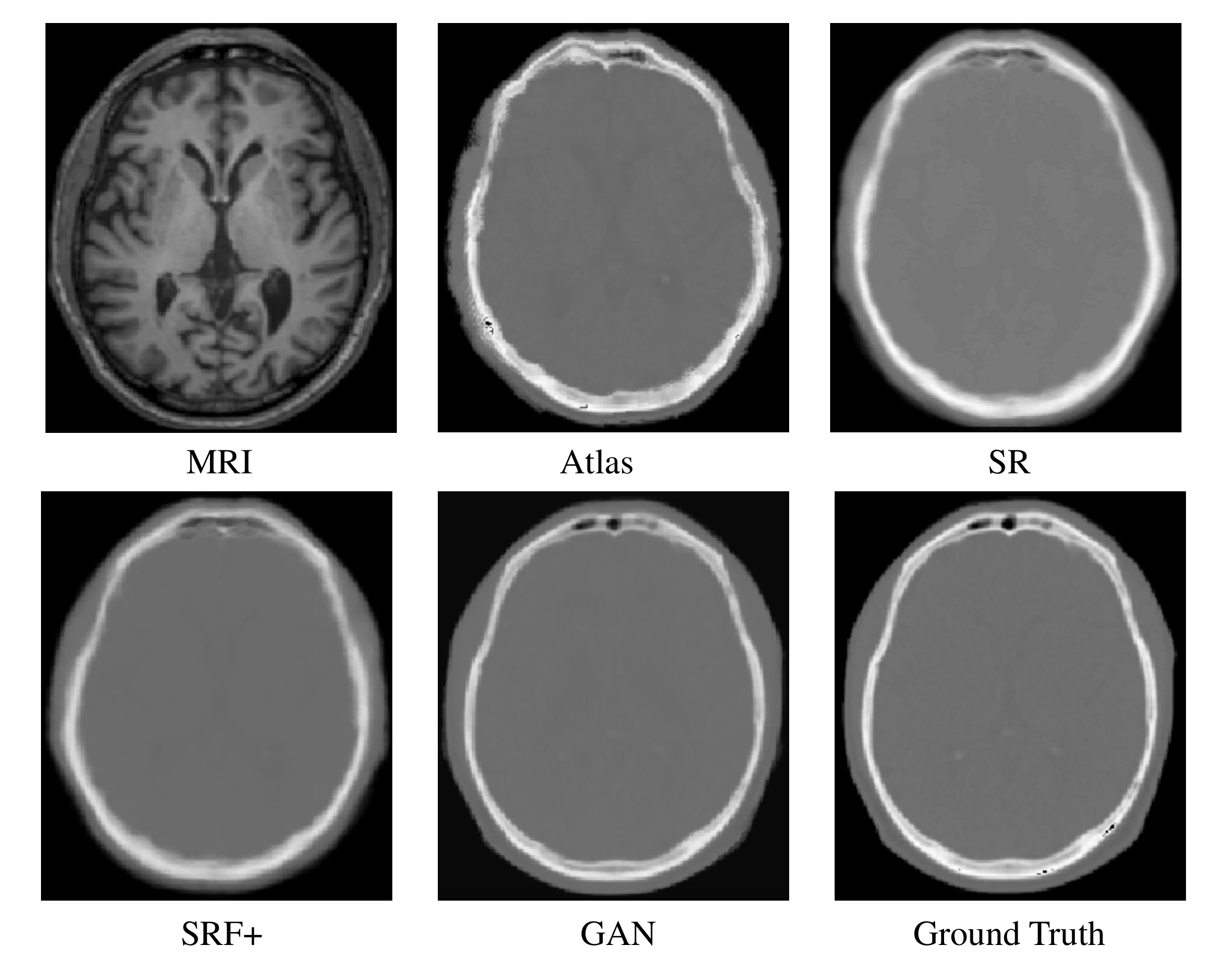}\\
  \caption{Visual comparison of original MR images, the estimated CT images by our method and other methods, and the ground-truth CT images on 1 typical subject on the brain dataset}
\label{fig:brain-res}
\vspace{-5pt}
\end{figure}

To qualitatively compare the estimated CT by different methods, we visualize the generated CT with the ground truth in Fig.~\ref{fig:brain-res}. We can see that the proposed algorithm can better preserve the continuity, coalition and smoothness in the prediction results, since it uses image gradient difference constraints in the image patch as discussed in Section ~\ref{subsec:gan}. Furthermore, the generated CT looks closer to the real CT compared to all others, and we argue that this is due to the adversarial training strategy which constrains the generated images to be so similar to the real ones that a even a complex discriminator cannot perform better than chance.

We also quantitatively compare the predicted results in Table ~\ref{tab:brain} in terms of PSNR and MAE. Our proposed method outperforms all other methods in both metrics, and it further demonstrates the advantage of our architecture.

\begin{table}[t]
\vspace{-5pt}
\caption{Average of MAE and PSNR on 16 subjects from the brain dataset by 4 different methods: Atlas, SR, SRF+, and Proposed}
\begin{minipage}{.5\linewidth}
\centering
\begin{tabular}{l ccc}
\hline
\textbf{MAE} & Mean(std.) & Med. \\
\hline
Atlas &171.5(35.7) & 170.2 \\
SR & 159.8(37.4) & 161.1 \\
SRF+ & 99.9(14.2)  & 97.6\\
Proposed & {\bf 92.5(13.9)}  &  {\bf 92.1}\\
\hline
\label{tab:brain-mae}
\end{tabular}
\end{minipage}\begin{minipage}{.5\linewidth}
\centering
\begin{tabular}{l ccc}
\hline
\textbf{PSNR} & Mean(std.) & Med. \\
\hline
Atlas  & 20.8(1.6)  & 20.6 \\
SR & 21.3(1.7) & 21.2\\
SRF+ & 26.3(1.4)  & 26.3\\
Proposed & {\bf 27.6(1.3)}  &  {\bf 27.6}\\
\hline
\label{tab:brain-psnr}
\end{tabular}
\end{minipage}
\label{tab:brain}
\vspace{-5pt}
\end{table}

\subsection{Experimental Results for Pelvic Dataset}
\begin{figure}[t!]
\centering
  \includegraphics[width=1\linewidth]{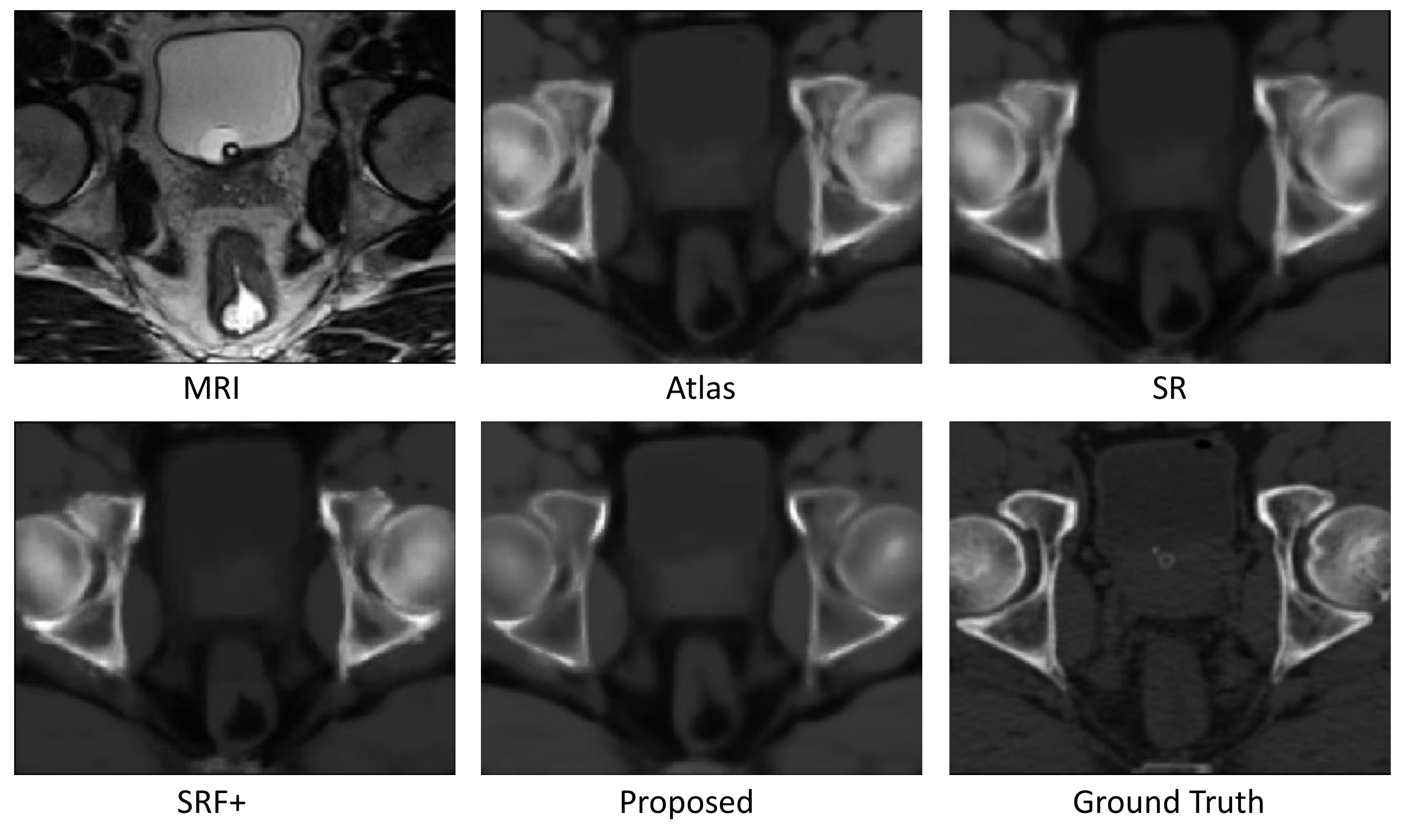}\\
  \caption{Visual comparison of original MR images, the estimated CT images by our method and other methods, and the ground-truth CT images on 1 typical subject on the pelvic dataset}
\label{fig:pros-res}
\end{figure}

The prediction results by the same methods used above but on the pelvic dataset are shown in Fig.~\ref{fig:pros-res}. It can be clearly seen that our results are consistent with the ground-truth CT. The quantitative results based on the same metrics used in the previous dataset are shown in Table ~\ref{tab:prostate}.

\begin{table}[t]
\caption{Average of MAE and PSNR on 22 subjects from the pelvic dataset by 4 different methods: Atlas, SR, SRF+, and Proposed}
\begin{minipage}{.5\linewidth}
\centering
\begin{tabular}{l ccc}
\hline
\textbf{MAE} & Mean(std.) & Med. \\
\hline
Atlas  & 66.1(6.9)  &  66.7 \\
SR &52.1(9.8) & 52.3 \\
SRF+ & 48.1(4.6)  & 48.3\\
Proposed & {\bf 39.0(4.6)}  &  {\bf 39.1}\\
\hline
\label{tab:prostate-mae}
\end{tabular}
\end{minipage}\begin{minipage}{.5\linewidth}
\centering
\begin{tabular}{l ccc}
\hline
\textbf{PSNR} & Mean(std.) & Med. \\
\hline
Atlas  & 29.0(2.1)  &  29.6 \\
SR &30.3(2.6) & 31.1  \\
SRF+ & 32.1(0.9)  & 31.8\\
Proposed & {\bf 34.1(1.0)}  &  {\bf 34.1}\\
\hline
\label{tab:prostate-psnr}
\end{tabular}
\end{minipage}
\label{tab:prostate}
\vspace{-5pt}
\end{table}

Quantitative results in Tables ~\ref{tab:prostate} show that our method outperforms the other methods in terms of both MAE and PSNR. Specifically, our method gives an average PSNR of 34.1, which is considerably higher than the 32.1 obtained by the state-of-the-art SRF+ method.

\section{Conclusions}
\label{sec:conclusion}
We have developed a 3D GAN model for estimating CT images from MRI images by directly taking MR image patches as input and CT patches as output. The performance is improved by the use of  ACM since the context of the GAN is effectively enlarged during the training process, which makes it context aware.
 We have applied this model to predict CT images from their corresponding MR images where the experiments demonstrate that our proposed method can significantly outperform three state-of-the-art methods, showing its suitability in regression tasks.  Although we only considered the task of CT image prediction, our proposed model can also be applied to other related tasks involving a generative process in medical image analysis such as super-resolution, image denoising and so on. 

%
%
\clearpage
\bibliographystyle{plain}
\bibliography{mrct}

\end{document}